\documentclass{article}

\usepackage[preprint]{neurips_2025}

\usepackage[utf8]{inputenc} 
\usepackage[T1]{fontenc}    
\usepackage{url}            
\usepackage{booktabs}       
\usepackage{amsfonts}       
\usepackage{nicefrac}       
\usepackage{microtype}      
\usepackage[dvipsnames]{xcolor}
\usepackage{graphicx}
\usepackage{amsmath}
\usepackage{multirow}

\usepackage{tikz}
\usepackage{caption}
\usepackage{subcaption}
\usepackage{makecell}
\usepackage{multicol}
\usepackage{colortbl}
\definecolor{myblue}{RGB}{230,245,255}
\definecolor{mygreen}{RGB}{1,109,101}

\definecolor{mydarkblue}{rgb}{0.68, 0.85, 1.0}
\definecolor{mydarkblue2}{rgb}{0,0.08,0.45}
\definecolor{mydarkblue3}{RGB}{151,204,255}
\definecolor{cvprblue}{rgb}{0.21,0.49,0.74}
\definecolor{oxfordblue}{RGB}{0,33,71}
\definecolor{oxfordroyalblue}{RGB}{29,66,166}

\usepackage[colorlinks=true,
    citecolor=cvprblue,
    filecolor=cvprblue,
    urlcolor=magenta]{hyperref}

\usepackage{cleveref}

\crefname{section}{Sec.}{Secs.}
\Crefname{section}{Section}{Sections}

\crefname{table}{Tab.}{Tabs.}
\Crefname{table}{Table}{Tables}

\crefname{figure}{Fig.}{Figs.}
\Crefname{figure}{Figure}{Figures}

\crefname{appendix}{App.}{Apps.}
\Crefname{appendix}{Appendix}{Appendices} 

\title{
\textcolor{oxfordroyalblue}{3DRS}: MLLMs Need {\color{oxfordroyalblue}{3D}}-Aware {\color{oxfordroyalblue}{R}}epresentation {\color{oxfordroyalblue}{S}}upervision for Scene Understanding
}

\author{
Xiaohu Huang\textsuperscript{1} \qquad
Jingjing Wu\textsuperscript{2} \qquad
\textbf{Qunyi Xie}\textsuperscript{2} \qquad
Kai Han\textsuperscript{1}\thanks{Corresponding author.} \\
\textsuperscript{1} Visual AI Lab, The University of Hong Kong \\
\textsuperscript{2} Department of Computer Vision Technology (VIS), Baidu Inc. \\
\texttt{huangxiaohu@connect.hku.hk, kaihanx@hku.hk}
}

\begin{document}

\maketitle

\vspace{-0.7cm}
\begin{abstract}
Recent advances in scene understanding have leveraged multimodal large language models (MLLMs) for 3D reasoning by capitalizing on their strong 2D pretraining. However, the lack of explicit 3D data during MLLM pretraining limits 3D representation capability. In this paper, we investigate the 3D-awareness of MLLMs by evaluating multi-view correspondence and reveal a strong positive correlation between the quality of 3D-aware representation and downstream task performance. Motivated by this, we propose {\color{oxfordroyalblue}{{3DRS}}}, a framework that enhances MLLM {\color{oxfordroyalblue}{\underline{3D}}} {\color{oxfordroyalblue}{\underline{R}}}epresentation learning by introducing {\color{oxfordroyalblue}{\underline{S}}}upervision from pretrained 3D foundation models. Our approach aligns MLLM visual features with rich 3D knowledge distilled from 3D models, effectively improving scene understanding. Extensive experiments across multiple benchmarks and MLLMs---including visual grounding, captioning, and question answering---demonstrate consistent performance gains. Project page: \url{https://visual-ai.github.io/3drs}
\end{abstract}
\vspace{-0.1cm}
\begin{figure}[h]
    \centering

    \begin{minipage}{0.7\textwidth}
        \raggedright
        {\scriptsize\textbf{(a) Overview of 3DRS.}}\\[0.2em]
        \centering
        \includegraphics[width=\textwidth]{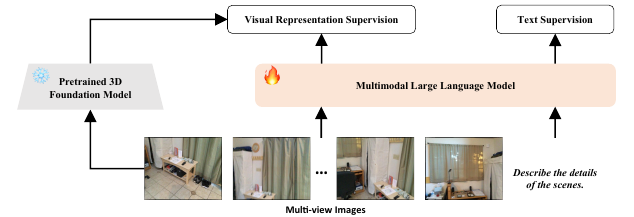}
    \end{minipage}
    \begin{center}
      \begin{tikzpicture}
        \draw[dashed, color=cyan, line width=0.8pt] (0,0) -- (0.8\textwidth,0);
      \end{tikzpicture}
    \end{center}
    \begin{minipage}{0.7\textwidth}
        \raggedright
        {\scriptsize\textbf{(b) Performance Improvement.}}\\[0.2em]
        \centering
        \includegraphics[width=\textwidth]{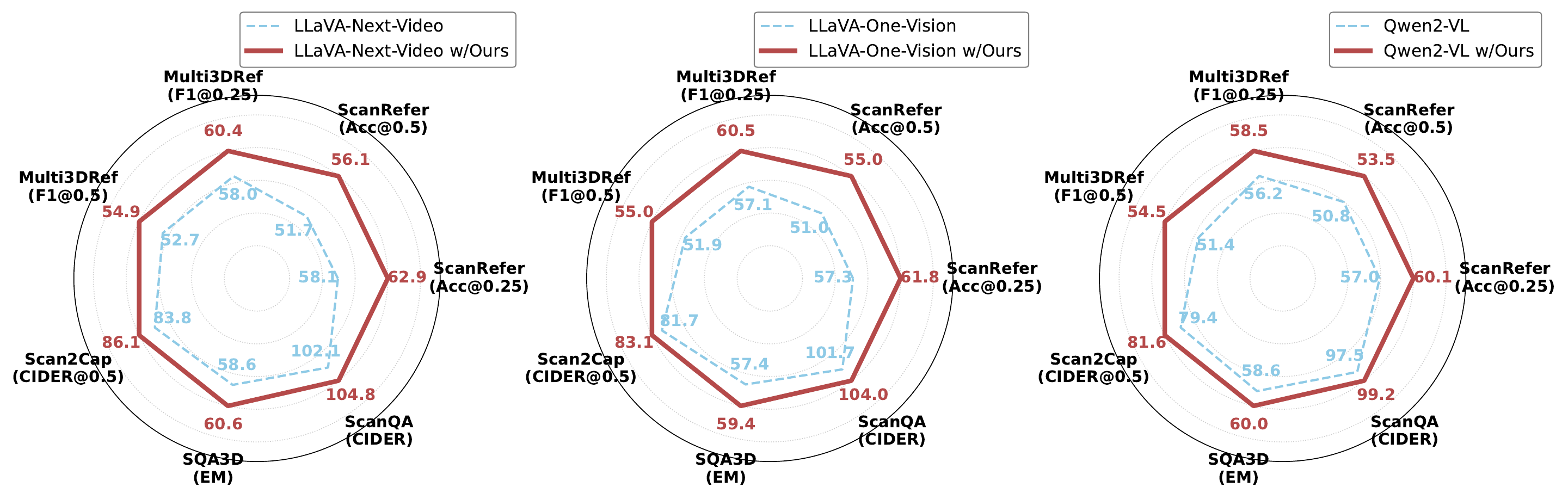}
    \end{minipage}

    \caption{
        \textbf{Enhancing 3D awareness of MLLMs to improve downstream performance.}
        (a) Besides the common text supervision for MLLMs, 3DRS adopts 3D foundation models to supervise 3D-aware visual representation learning in MLLMs.
        (b) Combined with 3DRS, we achieve consistent performance improvement across multiple MLLMs and benchmarks.
    }
    \label{fig:teaser}
\end{figure}

\section{Introduction}

Scene understanding serves as a cornerstone for interpreting 3D environments, enabling a wide range of critical applications ranging from robotic navigation to augmented reality. The recent emergence of large language models (LLMs)~\cite{qwen2, llama3, deepseekr1} has sparked innovative research aimed at endowing these models with scene comprehension capabilities. One major line of research~\cite{pointllm,chatscene, gpt4point, Point-bind, chatscene, minigpt, moretext, grounded-3dllm, 3dllava, ll3da, inst3d} utilizes point cloud encoders---either independently or in combination with multi-view images---to extract 3D representations, which are subsequently projected into a language-aligned space for LLMs. However, these approaches are constrained by the scarcity of paired 3D-text datasets, which impedes effective cross-modal feature alignment.

In response to this challenge, recent state-of-the-art methods \cite{llava3d, video3dllm, scenellm, 3d-llm, gpt4scene} have shifted towards leveraging multi-view images exclusively, drawing inspiration from the success of large-scale visual-language pretraining in multimodal LLMs (MLLMs) \cite{llava, llava-onevision, llavavideo, gpt-4, qwen2vl, team2023gemini}. These approaches aim to transfer 2D visual understanding to 3D scene comprehension by injecting 3D priors, such as 3D positional embeddings, into the models, thereby allowing MLLMs to capitalize on their extensive pretrained 2D knowledge for 3D interpretation.

Despite these advancements, genuine 3D scene understanding fundamentally requires models to capture intrinsic 3D attributes and spatial structures to comprehend scenes. The absence of explicit 3D data during MLLM pretraining reveals a significant gap, which motivates our core investigation centered around the following questions: (1) \textit{How can we evaluate the ability of MLLMs to learn 3D-aware representations?} (2) \textit{How does the quality of 3D feature learning influence downstream scene understanding performance?} (3) \textit{What methods can enhance 3D-aware representation learning within MLLM frameworks?} While several prior works~\cite{multiview,cua, probe3d,lexicon3d} have attempted to probe the 3D awareness of 2D vision foundation models, systematic investigation into 3D-aware representation learning in MLLMs remains largely unexplored. This gap is particularly crucial given the growing adoption of MLLMs in multimodal 3D understanding tasks. Our study aims to address this overlooked area and provide new insights into 3D representation learning within the MLLM paradigm.

For the first question, we conduct comprehensive experiments to evaluate the 3D awareness on three representative MLLMs, including LLaVA-Next-Video~\cite{llavavideo}, LLaVA-One-Vision~\cite{llava-onevision}, and Qwen2-VL~\cite{qwen2vl}, following the finetuing settings of Video-3D LLM~\cite{video3dllm}. Specifically, we assess 3D awareness via view equivariance, quantifying it by computing the feature similarity between corresponding pairs from the same 3D voxel across different views. This evaluation requires MLLMs to associate the same object across multiple views, thereby reflecting their capacity for 3D representation. Our analysis encompasses six datasets spanning tasks such as 3D grounding~\cite{scanrefer}, captioning~\cite{scan2cap}, and question answering~\cite{scanqa}.

To address the second question, we systematically analyze model performance across these datasets and observe that \textit{samples with higher correspondence scores---\textit{i.e.}, those exhibiting stronger 3D awareness---consistently lead to improved performance.} This finding demonstrates a strong positive correlation between the quality of 3D-aware representations and downstream scene understanding performance, highlighting the necessity of enhancing 3D feature learning in MLLMs.

In response to the third question and building upon our earlier findings, we first introduce a view equivalence supervision strategy for MLLMs, encouraging alignment between feature pairs corresponding to the same 3D location across different views (positive pairs) while discouraging similarity among unrelated pairs (negative pairs). While this approach results in some performance gains, the supervision provided by such handcrafted, single-task objectives is inherently limited for 3D learning.

In contrast, recent 3D foundation models such as VGGT~\cite{vggt} and FLARE~\cite{flare} are pretrained end-to-end on multi-view image sequences spanning a diverse set of 3D geometric tasks---including not only correspondence learning, but also depth estimation and camera parameter prediction. This comprehensive pretraining enables them to encode rich 3D properties within their features. Building on this, we propose a framework, 3DRS, that leverages these pretrained models by using their features as alignment targets for the visual outputs of MLLMs, thereby facilitating more effective 3D-aware representation learning. Unlike previous 3D MLLM approaches, in addition to traditional text token supervision, our framework employs explicit 3D-specific supervision directly on scene visual tokens. As demonstrated in our experiments (see \cref{fig:teaser}), incorporating this form of 3D supervision consistently improves performance across a range of MLLMs and benchmarks. Notably, our approach incurs no additional training overhead, since the supervisory features can be pre-extracted offline. We believe this design offers valuable new insights for applying 3D foundation models in scene understanding. The key contribution of this paper can be summarized as follows:

\begin{itemize}
    \item We conduct a systematic evaluation of the 3D-awareness of MLLMs using multi-view correspondence metrics, and observe a strong positive correlation between 3D-aware representation quality and downstream scene understanding performance across diverse tasks, datasets, and models.
    \item We propose a 3D-aware representation supervision framework that aligns MLLM visual features with those of a 3D geometry-pretrained model, enabling effective 3D feature learning.
    \item Extensive experiments demonstrate consistent performance improvements across multiple MLLMs and 3D scene understanding benchmarks, validating the effectiveness and generality of our approach.
\end{itemize}

\section{Method}
\label{sec:method}

\subsection{Investigating 3D-Aware Representation Learning in MLLMs}

\subsubsection{Preliminaries}
A MLLM typically consists of two main components: an image encoder $\mathcal{E}_\mathrm{img}$ and a text decoder $\mathcal{T}$. In this work, the input to our MLLM comprises a set of $N$ multi-view images $\mathcal{I} = \{I_1, I_2, \ldots, I_N\}$, each associated with per-pixel 3D coordinates $\mathcal{C} = \{\mathbf{C}_1, \mathbf{C}_2, \ldots, \mathbf{C}_N\}$, where $\mathbf{C}_i \in \mathbb{R}^{H \times W \times 3}$ for image $I_i$ of size $H \times W$. The 3D coordinates for each pixel are computed from the depth map and the corresponding camera intrinsic and extrinsic parameters; detailed formulas and procedures can be found in the \cref{appdix:corrdinate}.

The MLLM receives both multi-view images and language instructions as input. Internally, for each image $I_i$, the image encoder $\mathcal{E}_\mathrm{img}$ extracts visual features $\mathbf{F}_i \in \mathbb{R}^{H \times W \times d}$, where $d$ is the feature dimension. Following Video3DLLM~\cite{video3dllm}, we encode the per-pixel 3D coordinates via a positional encoding function $\phi(\cdot)$ and inject this information into the image features by addition:
\[
    \mathbf{F}_i^{3D} = \mathbf{F}_i + \phi(\mathbf{C}_i).
\]
This design allows the MLLM to inherit 2D perceptual knowledge from pretraining while equipping it with explicit 3D priors. 

During finetuning, the MLLM---which we denote as $f_\theta$---passes visual features $\{\mathbf{F}_i^{3D}\}_{i=1}^N$ with the instruction tokens to the text decoder for autoregressive text generation. After the processing of the text decoder, we refer to the final per-pixel visual embedding of pixel $p$ in image $I_i$ from LLM as $\mathbf{f}_i(p)$. The model is optimized by minimizing the standard cross-entropy loss:
\[
    \mathcal{L}_\mathrm{CE} = -\sum_{t=1}^{T} \log p_\theta(y_t \mid y_{<t}, \{I_i, \mathbf{C}_i\}_{i=1}^N, \text{instruction}),
\]
where $y_t$ is the $t$-th output token, and $p_\theta$ is the probability predicted by the model given all previous tokens and the multimodal context (\textit{i.e.}, images and instructions).


\subsubsection{Assessing 3D Feature Learning via Multi-View Correspondence}

Inspired by the crucial role of cross-view correspondences in 3D modeling~\cite{multipleviewgeometric}, we propose a correspondence-based evaluation framework. Multi-view correspondences are fundamental in 3D vision, serving as essential cues for core tasks such as ray retriangulation~\cite{multipleviewgeometric}, bundle adjustment~\cite{bundleadjustment}, and pose estimation~\cite{multiviewcategory}. They are also critical for downstream applications like instance recognition and retrieval~\cite{multiviewcategory,multishapereco,crossviewretrieval}. Therefore, we adopt multi-view correspondence analysis as a proxy to evaluate the 3D representations of MLLMs. This approach requires the model to accurately associate and align objects or regions that occupy the same position in 3D space across different viewpoints.

\noindent\textbf{Voxelization and correspondence pair construction.}
We first voxelize the 3D scene into a regular grid of voxels $\mathcal{V} = \{v_1, \ldots, v_M\}$. For each view $I_i$, given its per-pixel 3D coordinates $\mathbf{C}_i$, we assign every pixel's feature $\mathbf{f}_i(p)$ to a voxel according to its 3D position. Features from different views that fall into the same voxel $v_k$ are regarded as \emph{correspondence pairs}.

\noindent\textbf{Feature similarity and correspondence scores.}
Let $\mathcal{P}_k$ denote all correspondence feature pairs in voxel $v_k$, \textit{i.e.}, all pairs $(\mathbf{f}_i(p), \mathbf{f}_j(q))$ where both pixels $p$ and $q$ from images $I_i$ and $I_j$ are assigned to $v_k$ with $i \neq j$.
For any pair of visual features $(\mathbf{f}_a, \mathbf{f}_b)$ from the last layer of MLLM, feature similarity is measured by the cosine similarity:
\[
    S(\mathbf{f}_a, \mathbf{f}_b) = \frac{\mathbf{f}_a^\top \mathbf{f}_b}{\|\mathbf{f}_a\| \cdot \|\mathbf{f}_b\|}.
\]
For each sequence, we compute:
\begin{align*}
    \bar{S} &= \frac{1}{|\mathcal{P}|} \sum_{(\mathbf{f}_a, \mathbf{f}_b) \in \mathcal{P}} S(\mathbf{f}_a, \mathbf{f}_b),
\end{align*}
where $\bar{S}$ and $\mathcal{P}$ denote the \emph{correspondence score} for each sequence and all the correspondence pairs in this sequence.
A higher correspondence score indicates that the model produces more consistent features across views for the same 3D spatial location, reflecting stronger 3D-aware representation learning.

\subsubsection{Quality of 3D Feature \textbf{vs.} Downstream Task Performance.}

We evaluate three representative MLLMs, LLaVA-Next-Video, LLaVA-OneVision, and Qwen2-VL, on five diverse 3D scene understanding benchmarks, including visual grounding (Multi3DRefer, ScanRefer), captioning (Scan2Cap), and question answering (ScanQA, SQA3D). All benchmarks are based on multi-view RGBD sequences. The three MLLMs respectively emphasize video understanding, joint image-video reasoning, and advanced arbitrary-resolution visual encoding. 

\begin{figure}[htb]
    \centering
    \includegraphics[width=\linewidth]{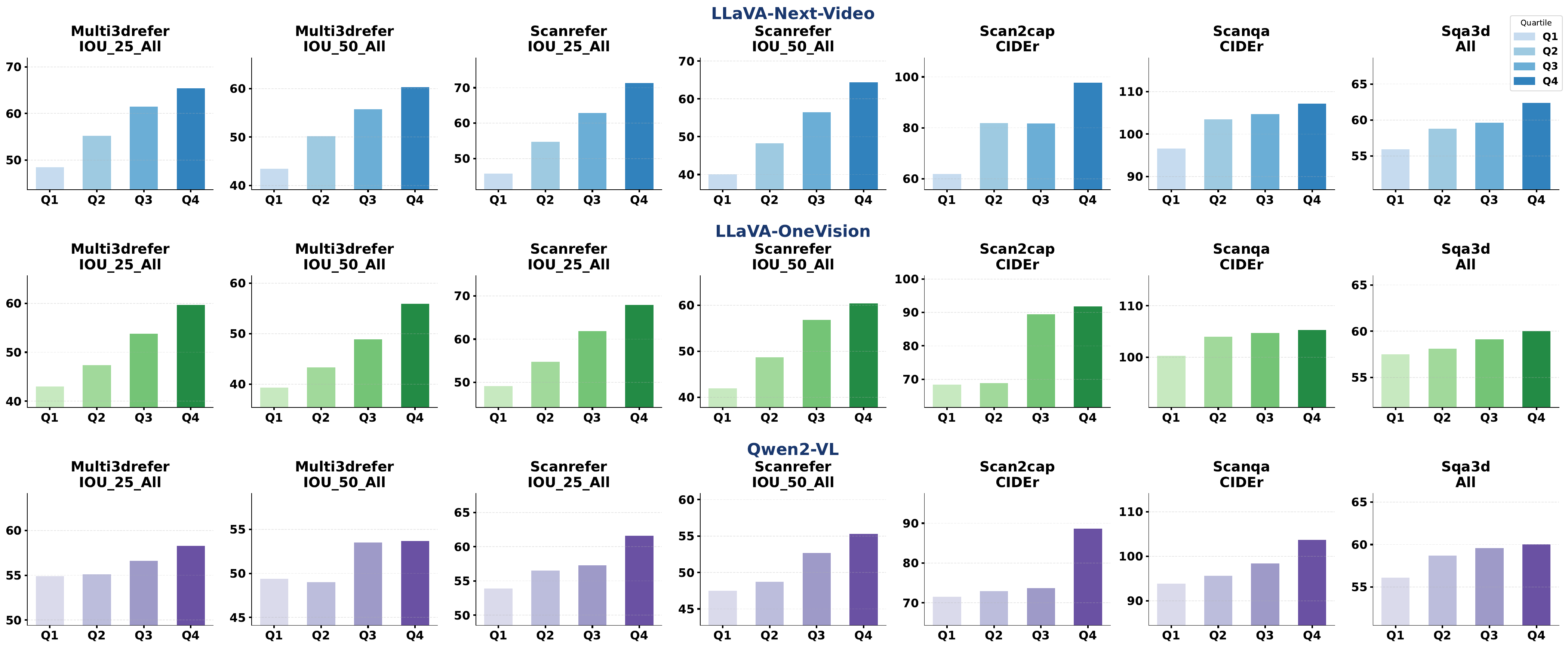}
    \caption{\textbf{Performance across correspondence score quartiles.} Model performance across correspondence score quartiles (Q1--Q4, lowest to highest) for each dataset. Samples were divided into quartiles by their correspondence scores. A clear trend is observed: model accuracy improves as the correspondence score increases.}
    \label{fig:correspondence_score}
\end{figure}

To analyze the relationship between 3D feature learning and downstream task performance, we sort samples within each dataset by their correspondence scores and divide them into four quartiles (Q1--Q4, lowest to highest). From~\cref{fig:correspondence_score}, we observe a clear trend: \textit{as the correspondence score increases, the model's performance on the downstream task consistently improves.} This strong positive correlation demonstrates the critical importance of 3D-aware representation quality for effective scene understanding in MLLMs.

These findings highlight the need for strategies to further enhance 3D-aware representation learning in MLLMs, which we address in the next section.

\subsection{Enhancing 3D-Aware Representation Learning in MLLMs}

\subsubsection{Correspondence-based 3D Supervision Loss}

Inspired by our correspondence-based evaluation, a straightforward approach is to directly supervise the MLLM's visual features to be consistent for matched 3D locations across views and dissimilar for mismatched locations.  We let $\mathcal{P}^+_k$ denote all positive feature pairs in voxel $v_k$, \textit{i.e.}, all pairs $(\mathbf{f}_i(p), \mathbf{f}_j(q))$ where pixels $p$ and $q$ from images $I_i$ and $I_j$ are assigned to $v_k$ with $i \neq j$. Similarly, $\mathcal{P}^-_k$ denotes negative pairs between $v_k$ and any other voxel $v_l$ ($l \neq k$). We supervise these objectives directly using a simple loss function by maximizing the feature similarity in $\mathcal{P}^+$ and minimizing that in $\mathcal{P}^-$:



\[
    \mathcal{L}_\mathrm{corr}^{+} = \frac{1}{|\mathcal{P}^+|} \sum_{(\mathbf{f}_a, \mathbf{f}_b) \in \mathcal{P}^+} \left[1 - S(\mathbf{f}_a, \mathbf{f}_b)\right],
\]
\[
    \mathcal{L}_\mathrm{corr}^{-} = \frac{1}{|\mathcal{P}^-|} \sum_{(\mathbf{f}_a, \mathbf{f}_b) \in \mathcal{P}^-} S(\mathbf{f}_a, \mathbf{f}_b).
\]

The overall correspondence loss is a weighted sum:
\[
    \mathcal{L}_\mathrm{corr} = \mathcal{L}_\mathrm{corr}^{+} + \mathcal{L}_\mathrm{corr}^{-}.
\]


By directly supervising positive pairs to be similar and negative pairs to be dissimilar, this correspondence loss encourages the model to learn multi-view 3D correspondences, thus enhancing the 3D-awareness of the learned representations. As will be shown in the experiments \cref{sec:exp}, supplementing the standard cross-entropy objective with $\mathcal{L}_\mathrm{corr}$ leads to improvements in downstream task performance. However, as this loss primarily targets view equivariance, the range of 3D properties captured remains limited, motivating the need for richer supervision.

\subsubsection{3D Foundation Model-Guided Feature Distillation}

To overcome the inherent limitations of single-task supervision, we further introduce a knowledge distillation framework, 3DRS, that leverages the rich 3D priors embedded in 3D foundation models, e.g, FLARE and VGGT. These models are pretrained on a wide array of 3D geometric tasks---including correspondence learning, camera parameter estimation, multi-view depth prediction, and dense point cloud reconstruction---which enables them to extract robust and highly 3D-aware visual features from multi-view image sequences.

\begin{figure}[t]  
    \centering
    \begin{subfigure}[t]{0.47\textwidth}
        \centering
        \includegraphics[width=\textwidth]{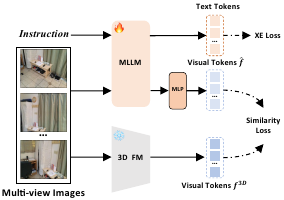}
        \caption{\textbf{Details of 3DRS.}}
        \label{fig:3drs_detail}
    \end{subfigure}
    \hspace{0.01\textwidth}
    \begin{subfigure}[t]{0.48\textwidth}
        \centering
        \includegraphics[width=\textwidth]{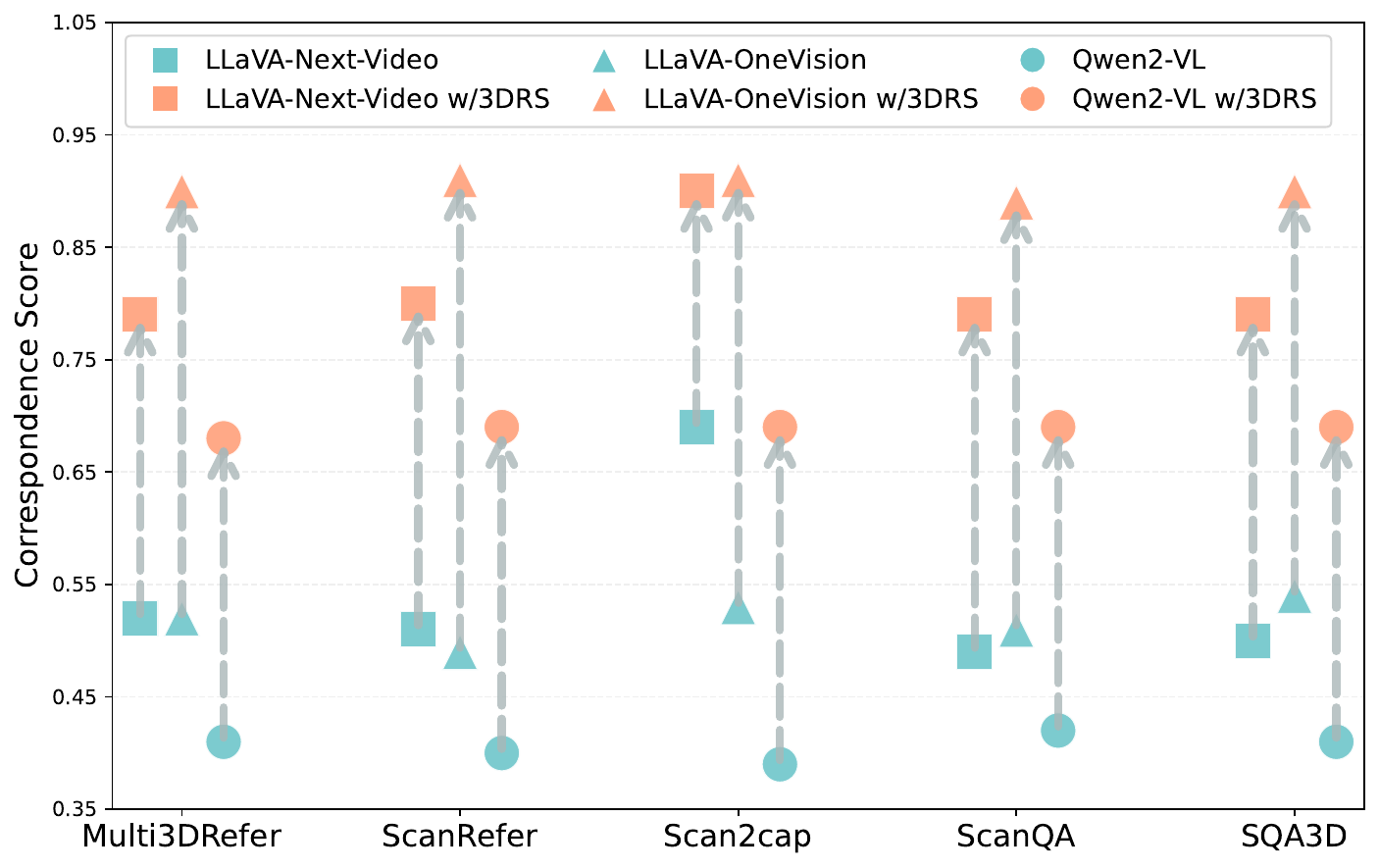}
        \caption{\textbf{Comparison of correspondence learning.}}
        \label{fig:3drs_result}
    \end{subfigure}
    \caption{
        (a) 3DRS uses a 3D foundation model to supervise the visual representation of the MLLM.
        (b) 3DRS effectively improves the correspondence learning for MLLMs.
    }
    \label{fig:3drs}
\end{figure}

\paragraph{Distillation target preparation.}
As shown in \cref{fig:3drs_detail}, given a set of multi-view images $\mathcal{I}$ for a scene, we first input them into a pretrained 3D foundation model $g$, which outputs a collection of per-pixel visual features $\{\mathbf{f}_i^{\mathrm{3D}}(p)\}$ for each image $I_i$ and pixel $p$. Since the spatial resolution of these features may differ from those of the MLLM outputs $\{\mathbf{f}_i(p)\}$, we apply 2D average pooling to the 3D foundation model's output to match the MLLM feature map size.


\paragraph{Feature alignment and loss.}
To align the MLLM's per-pixel visual features with the 3D foundation model, we first process each $\mathbf{f}_i(p)$ with a two-layer MLP (denoted as $\mathrm{MLP}_\mathrm{align}$) to ensure compatibility in feature dimension:
\[
    \hat{\mathbf{f}}_i(p) = \mathrm{MLP}_\mathrm{align}(\mathbf{f}_i(p)).
\]

We then employ a distillation loss based on euclidean similarity to maximize the alignment between the MLLM features $\hat{\mathbf{f}}_i(p)$ and the corresponding 3D foundation model features $\mathbf{f}_i^{\mathrm{3D}}(p)$:
\[
    \mathcal{L}_\mathrm{align} = - \frac{1}{N H W} \sum_{i=1}^{N} \sum_{p \in I_i} S\left( \hat{\mathbf{f}}_i(p), \mathbf{f}_i^{\mathrm{3D}}(p) \right),
\]
where $S(\cdot, \cdot)$ denotes cosine similarity, and the sum is calculated over all pixels and views in the batch.

\paragraph{Overall training objective.}
The final training objective for the MLLM combines the standard cross-entropy loss for text generation and the 3D foundation model distillation loss:
\[
    \mathcal{L}_\mathrm{total} = \mathcal{L}_\mathrm{CE} + \mathcal{L}_\mathrm{align}.
\]

This approach enables the MLLM to inherit comprehensive 3D knowledge from powerful geometry-pretrained models, facilitating the learning of richer and more robust 3D-aware representations. Importantly, the distillation targets from the 3D foundation model can be precomputed offline, introducing no additional overhead during MLLM fine-tuning. 

As illustrated in~\cref{fig:3drs_result}, we compare the correspondence scores before and after applying our 3DRS, where VGGT serves as the foundation model. The results consistently demonstrate that introducing 3DRS leads to substantial improvements in correspondence learning ability across all evaluated MLLMs and benchmarks. This proves the effectiveness of leveraging a pretrained 3D foundation model as a teacher model for enhancing 3D-aware representation learning in MLLMs. More comprehensive experimental results and analyses are detailed in \cref{sec:exp}.

\section{Experiments}
\label{sec:exp}

\subsection{Datasets and Evaluation Metrics}

\textbf{Datasets.}  We evaluate our approach on six benchmarks that collectively span key challenges in 3D scene understanding. ScanRefer~\cite{scanrefer} focuses on localizing objects using free-form language, while Multi3DRefer~\cite{multi3drefer} generalizes this to queries referencing zero, one, or multiple objects, better reflecting real-world ambiguity. Scan2Cap~\cite{scan2cap} addresses dense captioning by pairing detected objects in 3D scans with natural language descriptions. For question answering, ScanQA~\cite{scanqa} tasks models with answering open-ended questions grounded in 3D geometry and semantics, and SQA3D~\cite{sqa3d} goes further by requiring situated reasoning: agents must interpret their position and context to answer complex queries. All these datasets are sourced from the richly annotated ScanNet~\cite{scannet} corpus, and we follow standard validation and test splits as established in prior work~\cite{chatscene, llava3d, grounded-3dllm, video3dllm}. Besides, VSI-Bench~\cite{vsibench} is used to evaluate the performance on visual-based spatial understanding tasks, which are composed of numerical and multiple-choice questions. The statistics of training sets are detailed in the \cref{appdix:trainsets}.

\textbf{Evaluation metrics.} For ScanRefer, we report accuracy at IoU thresholds of 0.25 and 0.5 (Acc@0.25, Acc@0.5). Multi3DRefer uses F1 scores at matching IoU thresholds. Scan2Cap is evaluated by CIDEr and BLEU-4 scores at 0.5 IoU (C@0.5, B-4@0.5). ScanQA is assessed by CIDEr and exact match accuracy (C, EM), while SQA3D uses exact match accuracy as the metric.

\begin{table*}[t]
\centering
\small
\setlength{\tabcolsep}{1.5mm}
\caption{Performance comparison on 3D scene understanding benchmarks. Specialists are single-task methods, while generalists target multiple tasks. Bold denotes best performance.}
\resizebox{\linewidth}{!}{
\begin{tabular}{lcccccccccc}
\toprule
\multirow{2}{*}{Method}
& \multicolumn{2}{c}{ScanRefer} 
& \multicolumn{2}{c}{Multi3DRefer} 
& \multicolumn{2}{c}{Scan2Cap}  
& \multicolumn{2}{c}{ScanQA} 
& SQA3D \\ 
\cmidrule(lr){2-3} \cmidrule(lr){4-5} \cmidrule(lr){6-7} \cmidrule(lr){8-9} \cmidrule(lr){10-10}
& Acc@0.25 & Acc@0.5 & F1@0.25 & F1@0.5 & C@0.5 & B-4@0.5 & C & EM & EM \\
\midrule
\multicolumn{10}{l}{\textit{\textbf{Specialists}}} \\
ScanRefer~\cite{scanrefer}     & 37.3 & 24.3 & --   & --   & --   & --   & --   & --   & --   \\
MVT~\cite{mvt}                 & 40.8 & 33.3 & --   & --   & --   & --   & --   & --   & --   \\
3DVG-Trans~\cite{3dvg-trans}   & 45.9 & 34.5 & --   & --   & --   & --   & --   & --   & --   \\
ViL3DRel~\cite{vil3drel}       & 47.9 & 37.7 & --   & --   & --   & --   & --   & --   & --   \\
M3DRef-CLIP~\cite{multi3drefer}& 51.9 & 44.7 & 42.8 & --   & 38.4 & --   & --   & --   & --   \\
Scan2Cap~\cite{scan2cap}       & --   & --   & --   & --   & 35.2 & 22.4 & --   & --   & --   \\
ScanQA~\cite{scanqa}           & --   & --   & --   & --   & --   & --   & 64.9 & 21.1 & 47.2 \\
3D-VisTA~\cite{3dvista}        & 50.6 & 45.8 & --   & --   & 66.9 & 34.0 & 69.6 & 22.4 & 48.5 \\ 
\midrule
\multicolumn{10}{l}{\textit{\textbf{Generalists}}} \\
3D-LLM(Flamingo)~\cite{3d-llm}           & 21.2 & --   & --   & --   & --   & --   & 59.2 & 20.4 & --   \\
3D-LLM(BLIP2-flant5)~\cite{3d-llm}       & 30.3 & --   & --   & --   & --   & --   & 69.4 & 20.5 & --   \\
Chat-3D~\cite{chat3d}                    & --   & --   & --   & --   & --   & --   & 53.2 & --   & --   \\
Chat-3D v2~\cite{chatscene}              & 42.5 & 38.4 & 45.1 & 41.6 & 63.9 & 31.8 & 87.6 & --   & 54.7 \\
LL3DA~\cite{ll3da}                       & --   & --   & --   & --   & 62.9 & 36.0 & 76.8 & --   & --   \\
SceneLLM~\cite{scenellm}                 & --   & --   & --   & --   & --   & --   & 80.0 & 27.2 & 53.6 \\
LEO~\cite{leo}                           & --   & --   & --   & --   & 72.4 & 38.2 & 101.4& 21.5 & 50.0 \\
Grounded 3D-LLM~\cite{grounded-3dllm}    & 47.9 & 44.1 & 45.2 & 40.6 & 70.6 & 35.5 & 72.7 & --   & --   \\ 
PQ3D~\cite{pq3d}                         & 57.0 & 51.2 & --   & 50.1 & 80.3 & 36.0 & --   & --   & 47.1 \\
ChatScene~\cite{chatscene}               & 55.5 & 50.2 & 57.1 & 52.4 & 77.1 & 36.3 & 87.7 & 21.6 & 54.6 \\
LLaVA-3D~\cite{llava3d}                  & 54.1 & 42.4 & --   & --   & 79.2 & 41.1 & 91.7 & 27.0 & 55.6 \\
Inst3D-LLM~\cite{inst3d}                 & 57.8 & 51.6 & 58.3 & 53.5 & 79.7 & 38.3 & 88.6 & 24.6 & --   \\
3D-LLaVA~\cite{3dllava}                  & 51.2 & 40.6 & -- & -- & 78.8 & 36.9  & 92.6 & -- & 54.5 \\
Video-3D LLM~\cite{video3dllm}           & 58.1 & 51.7 & 58.0 & 52.7 & 83.8 & 41.3 & 102.1 & 30.1  & 58.6 \\
\rowcolor{myblue}
\textbf{3DRS}                            & \textbf{62.9} & \textbf{56.1} & \textbf{60.4} & \textbf{54.9} & \textbf{86.1} & \textbf{41.6} & \textbf{104.8} & \textbf{30.3} & \textbf{60.6} \\
\bottomrule
\end{tabular}
}
\label{tab:sota_performance}
\end{table*}

\subsection{Implementation Details}

Our experiments leverage several MLLMs, including LLaVA-Next-Video 7B~\cite{llavavideo}, LLaVA-OneVision 7B~\cite{llava-onevision}, and Qwen2-VL 7B~\cite{qwen2vl}. In addition to these baselines, we systematically compare the effect of using 2D versus 3D foundation models as teachers for MLLM finetuning. The 2D teacher models include DINOv2~\cite{dinov2}, MAE~\cite{mae}, and SigLIP ~\cite{siglip2}, while the 3D teacher models comprise FLARE~\cite{flare} and VGGT ~\cite{vggt}. Unless stated otherwise, we use LLaVA-Next-Video as the MLLM and VGGT as the representation teacher for our experiments.

For both training and inference, we uniformly sample 32 frames per scan to construct multi-view image sets. For evaluating the correspondence score, we use the voxel size of 0.1 for voxelization. All models are optimized using Adam, with a batch size of 16 and a warm-up ratio of 0.03. The learning rates are set to a maximum of $1\times10^{-5}$ for the language model and $2\times10^{-6}$ for the visual backbone during the warm-up period. During training for visual grounding and dense captioning, ground truth object regions are used as candidates, whereas during inference, we follow the procedure of~\cite{chatscene, leo, video3dllm} and employ Mask3D~\cite{mask3d} to generate object proposals. For LLaVA-Next-Video and LLaVA-OneVision, we finetune all model parameters. For Qwen2-VL, due to GPU memory constraints, we finetune only the projector and the LLM components. We use 8 H100 NVIDIA GPUs for all experiments.

\subsection{Comparison with State-of-the-Art Models}

\Cref{tab:sota_performance} presents a comprehensive comparison between our approach, task-specific specialist models---which require fine-tuning on individual datasets---and 3D generalist models that are capable of handling multiple tasks. Compared to specialist models, our approach achieves substantial performance improvements. This demonstrates the significant benefits brought by joint training and the LLM-based architecture, which contribute to superior generalization and feature integration compared to methods tailored for specific tasks. Furthermore, our method consistently outperforms 3D generalist approaches that utilize point clouds as input, such as LL3DA, Chat-3D, Grounded 3D-LLM, and 3D-LLaVA. Compared to Inst3D-LLM---which fuses multi-view images and point clouds---our approach also shows clear advantages, highlighting the strength of leveraging MLLMs as the backbone. Additionally, our method achieves considerable improvements over other MLLM-based methods, including LLaVA-3D and Video-3D LLM. These results collectively indicate that enhancing the 3D-awareness of MLLMs is highly effective for 3D scene understanding tasks, further validating the effectiveness of our proposed strategy.

\Cref{tab:vsi_bench_opensource} compares our method with leading proprietary APIs and open-source models on VSI-Bench~\cite{vsibench}, covering tasks such as object counting, spatial distance estimation, size reasoning, and sequence understanding. 3DRS achieves the best open-source results on most metrics—including object count, absolute distance, room size, and appearance oder—and remains competitive with proprietary models. These results demonstrate the strong spatial reasoning, generalization, and comprehensive scene understanding capabilities of our approach across diverse 3D vision tasks.

\begin{table*}[t]
\centering
\small
\setlength{\tabcolsep}{1.2mm}
\caption{Performance comparison on VSI-Bench.}
\resizebox{\linewidth}{!}{
\begin{tabular}{lccccccccc}
\toprule
Method & Avg. & Obj. Count & Abs. Distance & Obj. Size & Room Size & Rel. Dist. & Rel. Dir. & Route Plan & Appr. Order \\
\midrule
GPT-4o~\cite{gpt-4} & 34.0 & 46.2 & 5.3 & 43.8 & 38.2 & 37.0 & 41.3 & 31.5 & 28.5 \\
Gemini-1.5 Pro~\cite{team2023gemini} & 45.4 & 56.2 & 30.9 & \textbf{64.1} & 43.6 & \textbf{51.3} & \textbf{46.3} & \textbf{36.0} & 34.6 \\
\cmidrule(lr){1-10}
LongVA-7B~\cite{longva} & 29.2 & 38.0 & 22.2 & 33.1 & 43.3 & 25.4 & 15.7 & 33.1 & 17.7 \\
InternVL2-40B~\cite{internvl} & 37.0 & 41.3 & 26.2 & 48.2 & 27.5 & 47.6 & 32.7 & 27.8 & 44.7 \\
LLaVA-Video-7B~\cite{llavavideo} & 35.6 & 48.5 & 14.0 & 47.8 & 24.2 & 43.5 & 42.4 & 34.0 & 30.6 \\
LLaVA-Video-72B~\cite{llavavideo} & 40.9 & 48.9 & 22.8 & 57.4 & 35.3 & 42.4 & 36.7 & 35.0 & 48.6 \\
LLaVA-OneVision-7B~\cite{llava-onevision} & 32.4 & 47.7 & 20.7 & 47.4 & 42.5 & 35.2 & 29.4 & 24.4 \\
LLaVA-OneVision-72B~\cite{llava-onevision} & 40.2 & 43.5 & 23.9 & 57.6 & 37.5 & 42.5 & 39.9 & 32.5 & 44.6 \\
\rowcolor{myblue} \textbf{3DRS} & \textbf{45.9} & \textbf{68.7} & \textbf{34.8} & 53.6 & \textbf{56.6} & 40.9 & 43.2 & 30.4 & \textbf{39.2} \\
\bottomrule
\end{tabular}
}
\label{tab:vsi_bench_opensource}
\end{table*}

\begin{table}[h]
\centering
\small
\setlength{\tabcolsep}{1.5mm}
\renewcommand{\arraystretch}{1.1}
\caption{Performance comparison of 3DRS when using with different MLLMs.}
\resizebox{\linewidth}{!}{
\begin{tabular}{lcccccccccc}
\toprule
\multirow{2}{*}{Method}
& \multicolumn{2}{c}{ScanRefer} 
& \multicolumn{2}{c}{Multi3DRef} 
& \multicolumn{2}{c}{Scan2Cap}  
& \multicolumn{2}{c}{ScanQA} 
& SQA3D \\ 
\cmidrule(lr){2-3} \cmidrule(lr){4-5} \cmidrule(lr){6-7} \cmidrule(lr){8-9} \cmidrule(lr){10-10}
& Acc@0.25 & Acc@0.5 & F1@0.25 & F1@0.5 & B-4@0.5 & C@0.5 & C & EM & EM \\
\midrule
LLaVA-Next-Video~\cite{llavavideo} & 58.1 & 51.7 & 58.0 & 52.7 & 41.3 & 83.8 & 102.1 & 30.1 & 58.6 \\
\rowcolor{myblue}
\textbf{LLaVA-Next-Video w/ 3DRS} & \textbf{62.9} & \textbf{56.1} & \textbf{60.4} & \textbf{54.9} & \textbf{41.6} & \textbf{86.1} & \textbf{104.8} & \textbf{30.3} & \textbf{60.6} \\
LLaVA-OneVision~\cite{llava-onevision} & 57.3 & 51.0 & 57.1 & 51.9 & 40.4 & 81.7 & 101.7 & 29.0 & 57.4 \\
\rowcolor{myblue}
\textbf{LLaVA-OneVision w/ 3DRS} & \textbf{61.8} & \textbf{55.0} & \textbf{60.5} & \textbf{55.0} & \textbf{41.2} & \textbf{83.1} & \textbf{104.0} & \textbf{29.5} & \textbf{59.4} \\
Qwen2-VL~\cite{qwen2vl} & 57.0 & 50.8 & 56.2 & 51.4 & 39.5 & 79.4 & 97.5 & 28.7 & 58.6 \\
\rowcolor{myblue}
\textbf{Qwen2-VL w/ 3DRS} & \textbf{60.1} & \textbf{53.5} & \textbf{58.5} & \textbf{54.5} & \textbf{40.9} & \textbf{81.6} & \textbf{99.2} & \textbf{28.9} & \textbf{60.0} \\
\bottomrule
\end{tabular}
}
\label{tab:mllm_combinations}
\end{table}

\subsection{Diagnostic Study}
\textbf{Effectiveness with different MLLMs.} \Cref{tab:mllm_combinations} demonstrates that integrating 3DRS with different MLLMs—LLaVA-Next-Video, LLaVA-OneVision, and Qwen2-VL—consistently boosts performance across all evaluated benchmarks. For example, LLaVA-Next-Video w/ 3DRS improves ScanRefer Acc@0.25 from 58.1 to 62.9, and Multi3DRef F1@0.25 from 58.0 to 60.4. Similar gains are observed for LLaVA-OneVision and Qwen2-VL, where 3DRS brings improvements on every dataset and metric. These results highlight the general applicability of our approach and its effectiveness in enhancing 3D scene understanding for various MLLMs.

\textbf{Comparison between 2D and 3D foundation models.} \Cref{tab:foundation_supervisor} compares the performance of using 2D and 3D foundation models as representation supervisors. It is clear that 3D foundation models (FLARE and VGGT) outperform all 2D foundation models (MAE, Siglip2, Dinov2) across almost every metric. This performance gap can be attributed to the inherent difference in the prior knowledge captured by 2D and 3D foundation models. 3D models are pre-trained on large-scale 3D data and thus better capture geometric structure, spatial relationships, and depth information, which are critical for 3D scene understanding tasks. In contrast, 2D foundation models, trained on images, lack explicit 3D spatial priors and struggle to provide effective supervision for learning 3D-aware representations. This highlights the importance of 3D-specific foundation models for achieving strong results in downstream 3D tasks.

\textbf{Comparison of supervision signal.} \Cref{tab:supervision} shows that using correspondence loss for supervision brings improvements over the baseline, demonstrating the effectiveness of encouraging the model to learn multi-view correspondences. However, when 3D foundation model supervision is applied, the performance increases even further across all metrics. This indicates that 3D foundation models, with their rich 3D prior knowledge learned during pre-training, can more effectively enhance the 3D representation ability of MLLMs and yield greater gains for 3D understanding tasks.

\textbf{Comparison of supervision at different layers.} \Cref{tab:layer_supervision} examines the effect of applying 3D foundation model supervision at different layers of the network. The results reveal that supervision at deeper layers, especially the last layer, leads to the highest performance. This is likely because deeper layers are closer to the output and thus have a more direct impact on the final predictions. Additionally, these layers possess more parameters and a greater capacity to fit 3D features, which results in larger improvements on downstream tasks.

\begin{table}[h]
\centering
\small
\setlength{\tabcolsep}{1.5mm}
\renewcommand{\arraystretch}{1.1}
\caption{Ablation study on using different 2D/3D foundation models as the representation supervisor. Bold denotes the best in each group.}
\resizebox{\linewidth}{!}{
\begin{tabular}{lcccccccccc}
\toprule
\multirow{2}{*}{Representation Supervisor}
& \multicolumn{2}{c}{ScanRefer} 
& \multicolumn{2}{c}{Multi3DRef} 
& \multicolumn{2}{c}{Scan2Cap}  
& \multicolumn{2}{c}{ScanQA} 
& SQA3D \\
\cmidrule(lr){2-3} \cmidrule(lr){4-5} \cmidrule(lr){6-7} \cmidrule(lr){8-9} \cmidrule(lr){10-10}
& Acc@0.25 & Acc@0.5 & F1@0.25 & F1@0.5 & C@0.5 & B-4@0.5 & C & EM & EM \\
\midrule
Baseline & 58.1 & 51.7 & 58.0 & 52.7 & 83.8 & 41.3 & 102.1 & 30.1 & 58.6 \\
\midrule
\multicolumn{10}{l}{\textit{2D Foundation Models}} \\
Siglip2~\cite{siglip2} & 58.2 & 52.9  & 59.7 & 53.1 & 81.7  & 40.2  & 100.2 & 29.1 & 59.4 \\
MAE~\cite{mae}  &  59.1 & \textbf{53.7} & \textbf{60.0} & \textbf{53.7} & \textbf{82.8} & \textbf{40.4} & 102.5 & 29.5 & 59.2 \\
Dinov2~\cite{dinov2} & \textbf{59.8}  & 53.3  & 58.5 & 53.5  & 80.3 & 39.3 & \textbf{103.5} & \textbf{29.6} & \textbf{60.1} \\
\midrule
\multicolumn{10}{l}{\textit{3D Foundation Models}} \\
FLARE~\cite{flare}    & 62.1 & 55.7 & 59.8 & 54.8 & \textbf{86.6} & \textbf{42.5} & 104.4 & 30.1 & 60.1 \\
\rowcolor{myblue} VGGT~\cite{vggt}      & \textbf{62.9} & \textbf{56.1} & \textbf{60.4} & \textbf{54.9} & 86.1 & 41.6 & \textbf{104.8} & \textbf{30.3} & \textbf{60.6} \\
\bottomrule
\end{tabular}
}

\label{tab:foundation_supervisor}
\end{table}

\begin{table*}[t]
\centering
\begin{minipage}{0.5\linewidth}
\centering
\scriptsize
\setlength{\tabcolsep}{1mm}
\renewcommand{\arraystretch}{1.1}
\caption{Comparison of different supervision strategies.}
\begin{tabular}{lcccc}
\toprule
Method & \multicolumn{2}{c}{Multi3DRef} & \multicolumn{2}{c}{ScanRefer} \\
\cmidrule(lr){2-3} \cmidrule(lr){4-5}
& Acc@0.25 & Acc@0.5 & F1@0.25 & F1@0.5\\
\midrule
Baseline                  & 58.0 & 52.7 & 58.1 & 51.7 \\ 
w/ Correspondence Loss       & 60.1 & 53.3 & 59.1 & 53.7 \\
\rowcolor{myblue} w/ 3D Supervision      & \textbf{62.9} & \textbf{56.1} & \textbf{60.4} & \textbf{54.9} \\
\bottomrule
\end{tabular}
\label{tab:supervision}
\end{minipage}
\hfill
\begin{minipage}{0.48\linewidth}
\centering
\scriptsize
\setlength{\tabcolsep}{1mm}
\caption{3D foundation model supervision at different layers.}
\begin{tabular}{lcccc}
\toprule
Layer & \multicolumn{2}{c}{Multi3DRef} & \multicolumn{2}{c}{ScanRefer} \\
\cmidrule(lr){2-3} \cmidrule(lr){4-5}
 & Acc@0.25 & Acc@0.5 & F1@0.25 & F1@0.5 \\
\midrule
\rowcolor{myblue} Last Layer   &\textbf{62.9} & \textbf{56.1} & \textbf{60.4} & \textbf{54.9} \\
3rd Last Layer  & 61.7  & 54.9  & 59.7  & 54.3  \\
5th Last Layer  &  61.4  & 54.8  & 59.3  & 54.0  \\
10th Last Layer & 59.1 & 53.6 & 53.3 & 53.8 \\
\bottomrule
\end{tabular}
\label{tab:layer_supervision}
\end{minipage}
\end{table*}

\section{Related Work}

\subsection{Scene Understanding with Large Language Models}

LLMs, owing to their strong reasoning capabilities and remarkable success in 2D image understanding, have been widely applied to scene understanding tasks. Early works such as PointLLM~\cite{pointllm}, Point-Bind~\cite{Point-bind}, GPT4Point~\cite{gpt4point}, MiniGPT-3D~\cite{minigpt}, and Chat-3D~\cite{chat3d} leverage the alignment between point cloud and text features to facilitate 3D scene comprehension. Building on this foundation, methods like Grounded 3D-LLM~\cite{grounded-3dllm}, LL3DA~\cite{ll3da}, 3D-LLaVA~\cite{3dllava}, and Inst3D-LLM~\cite{inst3d} design more advanced cross-modal modules to better fuse multi-modal features, thereby enhancing scene representations. Furthermore, Chat-Scene~\cite{chatscene} and Inst3D-LLM~\cite{inst3d} exploit the complementary nature of 2D and 3D features to further boost scene understanding.

Some recent approaches, such as 3D-LLM~\cite{3d-llm} and Scene LLM~\cite{scenellm}, employ multi-view inputs and introduce 3D priors to transform 2D representations into a 3D-aware format. Thanks to pre-training on large-scale image-text datasets, methods based on MLLMs are gaining increasing popularity in the field of scene understanding. For instance, LLaVA-3D~\cite{llava3d} takes multi-view images as input and utilizes voxelization to reduce the dimensionality of representations, thus lowering computational costs while leveraging the strengths of MLLMs. However, many MLLMs require specially structured inputs, making them incompatible with certain approaches. Video 3D-LLM~\cite{video3dllm} and GPT4Scene~\cite{gpt4scene} more naturally inherit the MLLM pipeline by introducing 3D priors---such as positional embeddings or spatial markers---enabling the model to better comprehend 3D scene content. 

Our work follows this line of MLLM-based scene understanding, aiming to probe the 3D-awareness of MLLMs and analyze their relationships with downstream tasks. In particular, we demonstrate that introducing guidance from 3D foundation models can effectively enhance the representational capability of MLLMs for 3D scene understanding.

\subsection{3D-Awareness in Vision Models}


Several studies have investigated 3D-awareness; however, most prior work has focused on pure vision models rather than MLLMs, and primarily leveraged 2D foundation models instead of 3D ones. For example, FiT3D~\cite{yue2024improving}, Probe3D~\cite{probe3d}, and Lexicon3D~\cite{lexicon3d} empirically analyze the 3D-awareness of visual foundation models. CUA-O3D~\cite{cua} proposes integrating multiple 2D foundation models for 3D scene understanding, while Yang et al.\cite{multiview} evaluates and enhances the 3D-awareness of ViT-based models in various downstream tasks. Some previous 3D detection works~\cite{chen2022bevdistill,wang2023distillbev,bang2024radardistill} have focused on improving 3D representations for pure vision models or CLIP-style vision-language models (VLMs), primarily aiming to enhance geometric understanding and spatial localization within unimodal or early multimodal frameworks. In addition, several studies on scene understanding~\cite{3d2ddis,pri3d,peng2023openscene} have investigated various strategies for distilling 3D representations, such as transferring knowledge from 3D models to 2D networks or promoting cross-modal alignment. However, these efforts have not addressed the unique challenges presented by MLLMs, which require a more holistic integration of visual, linguistic, and spatial information. As a result, the potential of distilling 3D awareness into MLLMs for richer and more generalizable scene understanding remains largely unexplored.

In contrast, our work specifically targets the 3D-awareness of MLLMs. Rather than enhancing 3D feature learning via 2D foundation models, we introduce 3D foundation models as supervisors to directly guide and improve the 3D representation capabilities of MLLMs.

\section{Conclusion}


In this paper, we present a comprehensive study of the 3D representation capabilities of multi-modal large language models (MLLMs) in the context of scene understanding. While most existing research has centered on leveraging 2D foundation models to improve visual reasoning in MLLMs, the role and utility of 3D foundation models in this setting remain largely unexplored. To bridge this gap, we propose 3DRS, a novel framework that introduces direct 3D-aware supervision to MLLMs by leveraging pretrained 3D foundation models as teachers. Our approach enables MLLMs to acquire richer geometric and spatial representations, facilitating more accurate and robust understanding of complex 3D scenes. Through extensive experiments on diverse 3D scene understanding benchmarks, we demonstrate that 3DRS consistently improves performance across a variety of tasks, such as object localization, spatial reasoning, and 3D question answering. These results highlight the unique advantages and significant potential of integrating 3D foundation models for advancing multimodal scene understanding.


\section{Limitation}
\label{appdix:limitation}
While our paper aims to enhance the 3D-awareness of MLLMs, the relatively limited size of the dataset used for finetuning—especially when compared to that used during the MLLM pretraining stage—may restrict the full realization of our approach’s potential. Consequently, the improvements demonstrated in this work may only represent an initial step toward more robust 3D understanding. A promising direction for future research is to incorporate 3D-awareness learning into the pretraining stage of MLLMs, which could lead to fundamentally stronger models with deeper 3D comprehension. Besides,  due to the distillation-based nature of our approach, the performance of our method is upper-bounded by the quality of the teacher 3D foundation model. Any limitations or failure modes of the teacher—such as inaccurate correspondence, erroneous depth estimation, or incomplete geometric representations—can be propagated to the student MLLM and may potentially mislead it during training. While our experiments demonstrate consistent improvements over strong baselines, it is possible that errors or biases in the teacher's predictions can negatively impact the downstream 3D reasoning abilities of the student model. We believe that, as 3D foundation models continue to rapidly advance, this limitation becomes less pronounced over time. 

\newpage

\section*{Acknowledgments}

This work is supported by Hong Kong Research
Grant Council - General Research Fund (Grant No. 17213825).
We would like to thank Weining Ren for the valuable and insightful discussions.


{\small
\bibliographystyle{plain}
\bibliography{main}
\label{sec:ref}
}







\appendix
\newpage
\section{Technical Appendices and Supplementary Material}
\subsection{World Coordinate Computation}
\label{appdix:corrdinate}

\begin{table}[b]
    \centering
    \caption{Performance comparison on the validation set of ScanRefer \cite{scanrefer}.
    ``Unique'' and ``Multiple'' depends on whether there are other objects of the same class as the target object.
    } 
    {\small
    \resizebox{\textwidth}{!}{
    \begin{tabular}{lcccccc}
        \toprule
        \multirow{2}{*}{Method} & \multicolumn{2}{c}{Unique} & \multicolumn{2}{c}{Multiple} & \multicolumn{2}{c}{Overall} \\
         & Acc@0.25 & Acc@0.5 & Acc@0.25 & Acc@0.5 & Acc@0.25 & Acc@0.5 \\
         \midrule
        ScanRefer~\cite{scanrefer} & 76.3 & 53.5 & 32.7 & 21.1 & 41.2 & 27.4 \\
        MVT~\cite{mvt} & 77.7 & 66.4 & 31.9 & 25.3 & 40.8 & 33.3 \\
        3DVG-Transformer~\cite{3dvg-trans} & 81.9 & 60.6 & 39.3 & 28.4 & 47.6 & 34.7 \\
        ViL3DRel~\cite{vil3drel} & 81.6 & 68.6 & 40.3 & 30.7 & 47.9 & 37.7 \\
        3DJCG~\cite{3djcg} & 83.5 & 64.3 & 41.4 & 30.8 & 49.6 & 37.3 \\
        D3Net~\cite{d3net} & -- & 72.0 & -- & 30.1 & -- & 37.9 \\
        M3DRef-CLIP~\cite{multi3drefer} & 85.3 & 77.2 & 43.8 & 36.8 & 51.9 & 44.7 \\
        3D-VisTA~\cite{3dvista} & 81.6 & 75.1 & 43.7 & 39.1 & 50.6 & 45.8 \\
        3D-LLM (Flamingo) \cite{3d-llm} & -- & -- & -- & -- & 21.2 & -- \\
        3D-LLM (BLIP2-flant5) \cite{3d-llm} & -- & -- & -- & -- & 30.3 & -- \\
        Grounded 3D-LLM \cite{grounded-3dllm} & -- & -- & -- & -- & 47.9 & 44.1 \\
        PQ3D \cite{pq3d} & 86.7 & 78.3 & 51.5 & 46.2 & 57.0 & 51.2 \\
        ChatScene \cite{chatscene} & {89.6} & {82.5} & 47.8 & 42.9 & 55.5 & 50.2 \\
        LLaVA-3D \cite{llava3d} & -- & -- & -- & -- & 54.1 & 42.2\\
        Video 3D-LLM~\cite{video3dllm} & 88.0 & 78.3 & 50.9 & 45.3 & 58.1 & 51.7  \\
        \rowcolor{myblue}
        \textbf{3DRS (Ours)} & 87.4 & 77.9 & {57.0} & {50.8} & {62.9} & {56.1} \\
        \bottomrule
    \end{tabular}
    }
    }
    \label{tab:scanrefer}
\end{table}

Given a set of $N$ images $\mathcal{I} = \{I_1, I_2, \ldots, I_N\}$, each image $I_i$ is paired with its depth map $D_i \in \mathbb{R}^{H \times W}$, camera intrinsic matrix $K_i \in \mathbb{R}^{3 \times 3}$, and camera-to-world extrinsic matrix $T_i \in \mathbb{R}^{4 \times 4}$. For a pixel at $(u, v)$ in image $I_i$, the corresponding 3D coordinate in the global coordinate system, denoted as $\mathbf{C}_i(u, v) \in \mathbb{R}^3$, is computed as:

\begin{equation}
\begin{bmatrix}
\mathbf{C}_i(u, v) \\
1
\end{bmatrix}
=
T_i
\begin{bmatrix}
D_i(u, v) \cdot K_i^{-1}
\begin{bmatrix}
u \\
v \\
1
\end{bmatrix} \\
1
\end{bmatrix}
\end{equation}

Repeating this process for all pixels yields the per-pixel 3D coordinate map $\mathbf{C}_i \in \mathbb{R}^{H \times W \times 3}$ for each image $I_i$. The complete set of coordinate maps is denoted as $\mathcal{C} = \{\mathbf{C}_1, \mathbf{C}_2, \ldots, \mathbf{C}_N\}$.

\subsection{Datsests for Training}
\label{appdix:trainsets}
For model fine-tuning, we utilize a collection of well-established 3D vision-language datasets. Specifically, we follow the model finetuning settings of Video-3D LLM~\cite{video3dllm} by using the validation splits of ScanRefer, Multi3DRefer, Scan2Cap, and ScanQA, as well as the test split of SQA3D. 
Across these datasets, the number of data samples varies significantly: ScanRefer and Scan2Cap each provide 36,665 samples, while Multi3DRefer offers 43,838 entries. ScanQA contains 26,515 instances, and SQA3D is the largest with 79,445 samples. Most datasets are derived from 562 unique scans, except SQA3D, which includes 518 scans. We further report the average lengths of questions and answers for each dataset. For example, question lengths range from approximately 13 to 38 words, with Scan2Cap and ScanQA also providing answer texts, averaging 17.9 and 2.4 words in length, respectively. In SQA3D, the average question and answer lengths are 37.8 and 1.1 words. For the evaluation on VSI-Bench, we use the pre-training data from VG-LLM~\cite{vg-llm}.

\subsection{Detailed Comparison}
In this section, we provide a detailed comparison with other methods using all metrics across 5 benchmarks.

\textbf{Scanrefer.} \cref{tab:scanrefer} shows that our method 3DRS achieves the best overall performance on the ScanRefer validation set, especially in the challenging “Multiple” scenario where precise target discrimination is required. These results demonstrate that 3DRS effectively leverages multi-view images for robust spatial understanding and accurate object localization.

\textbf{Multi3DRefer.} In \cref{tab:multi3drefer}, 3DRS achieves the best overall results on the Multi3DRefer validation set, with top F1 scores in both standard and challenging scenarios. Our method consistently outperforms previous approaches, especially in the difficult zero-target and distractor settings, demonstrating superior robustness and spatial understanding.

\begin{table}[t]
    \centering
    \caption{{Performance comparison on the validation set of Multi3DRefer \cite{multi3drefer}. ZT: zero-target, ST: single-target, MT: multi-target, D: distractor.}}
    \small
    \resizebox{\textwidth}{!}{
    \begin{tabular}{lcccccccccc}
        \toprule
        \multirow{2}{*}{Method}  & ZT w/o D & ZT w/ D & \multicolumn{2}{c}{ST w/o D} & \multicolumn{2}{c}{ST w/ D} &\multicolumn{2}{c}{MT} & \multicolumn{2}{c}{ALL} \\
         & F1 & F1 & F1@0.25 & F1@0.5 & F1@0.25 & F1@0.5 & F1@0.25 & F1@0.5 & F1@0.25 & F1@0.5 \\
         \midrule
        M3DRef-CLIP~\cite{multi3drefer} & 81.8 & 39.4 & 53.5 & 47.8 & 34.6 & 30.6 & 43.6 & 37.9 & 42.8 & 38.4 \\
        D3Net~\cite{d3net}  & 81.6 & 32.5 & -- & 38.6 & -- & 23.3 & -- & 35.0 & -- & 32.2 \\
        3DJCG~\cite{3djcg} & 94.1 & 66.9 & -- & 26.0 & -- & 16.7 & -- & 26.2 & -- & 26.6 \\
        Grounded 3D-LLM \cite{grounded-3dllm} & -- & -- & -- & -- & -- & -- & -- & -- & 45.2 & 40.6 \\
        PQ3D \cite{pq3d} & 85.4 & 57.7 & -- & 68.5 & -- & 43.6 & -- & 40.9 & -- & 50.1 \\
        ChatScene \cite{chatscene}  & 90.3 & 62.6 & {82.9} & {75.9} & 49.1 & 44.5 & {45.7} & {41.1} & 57.1 & 52.4 \\
        Video 3D-LLM~\cite{video3dllm} & 94.7 & 78.5 & 82.6 & 73.4 & 52.1 & 47.2 & 40.8 & 35.7 & 58.0 & 52.7 \\
        \rowcolor{myblue}
        \textbf{3DRS (Ours)} & {95.6} & {79.4} & 79.6 & 71.4 & {57.0} & {51.3} & 43.0 & 37.8 & {60.4} & {54.9} \\
        \bottomrule
    \end{tabular}
    }
    \label{tab:multi3drefer}
\end{table}

\textbf{ScanQA.} In \cref{tab:scanqa}, 3DRS achieves the best performance on the ScanQA validation set across almost all metrics, including EM, BLEU scores, METEOR, and CIDEr, demonstrating its strong effectiveness for 3D question answering.

\begin{table}[t]
    \centering
    \caption{Performance comparison on the validation set of ScanQA \cite{scanqa}. EM indicates exact match accuracy, and B-1, B-2, B-3, B-4 denote BLEU-1, -2, -3, -4, respectively.}
    {\small
    \resizebox{\textwidth}{!}{
    \begin{tabular}{lcccccccc}
        \toprule
        Method & EM & B-1 & B-2 & B-3 & B-4 & ROUGE-L & METEOR & CIDEr \\
        \midrule
        ScanQA~\cite{scanqa} & 21.05 & 30.24 & 20.40 & 15.11 & 10.08 & 33.33 & 13.14 & 64.86  \\
        3D-VisTA \cite{3dvista} & 22.40 & -- & -- & -- & 10.40 & 35.70 & 13.90 & 69.60 \\
        Oryx-34B \cite{oryx} & -- & 38.00 & 24.60 & -- & -- & 37.30 & 15.00 & 72.30 \\
        LLaVA-Video-7B \cite{llavavideo} & -- & 39.71 & 26.57 & 9.33 & 3.09 & 44.62 & 17.72 & 88.70 \\
        3D-LLM (Flamingo) \cite{3d-llm} & 20.40 & 30.30 & 17.80 & 12.00 & 7.20 & 32.30 & 12.20 & 59.20 \\
        3D-LLM (BLIP2-flant5)~\cite{3d-llm} & 20.50 & 39.30 & 25.20 & 18.40 & 12.00 & 35.70 & 14.50 & 69.40  \\
        Chat-3D \cite{chat3d} & -- & 29.10 & -- & -- & 6.40 & 28.50 & 11.90 & 53.20 \\
        NaviLLM \cite{navillm} & 23.00 & -- & -- & -- & 12.50 & 38.40 & 15.40 & 75.90 \\
        LL3DA~\cite{ll3da} & -- & -- & -- & -- & 13.53 & 37.31 & 15.88 & 76.79 \\
        Scene-LLM~\cite{scenellm} & 27.20 & 43.60 & 26.80 & 19.10 & 12.00 & 40.00 & 16.60 & 80.00 \\
        LEO~\cite{leo} & -- & -- & -- & -- & 11.50 & 39.30 & 16.20 & 80.00 \\
        Grounded 3D-LLM \cite{grounded-3dllm} & -- & -- & -- & -- & 13.40 & -- & -- & 72.70\\
        ChatScene \cite{chatscene} & 21.62 & 43.20 & 29.06 & 20.57 & 14.31 & 41.56 & 18.00 & 87.70 \\
        LLaVA-3D \cite{llava3d} & 27.00 & -- & -- & -- & 14.50 & {50.10} & {20.70} &  91.70 \\
        Video 3D-LLM~\cite{video3dllm} & {30.10} & {47.05} & {31.70} & {22.83} & 16.17 & 49.02 & 19.84 & {102.06} \\
        \rowcolor{myblue}
        \textbf{3DRS (Ours)} & {30.30} & {48.37} & {32.67} & {23.79} & {17.22} & {49.82} & {20.47} & {104.78} \\
        \bottomrule
    \end{tabular}
    }
    }
    \label{tab:scanqa}
\end{table}

\textbf{SQA3D.} In \cref{tab:sqa3d}, 3DRS achieves the highest scores on the SQA3D test set, outperforming all previous approaches on almost every question type as well as in the overall average, which demonstrates its superior capability for 3D question answering across diverse scenarios.

\textbf{Scan2cap.} In \cref{tab:scan2cap}, 3DRS achieves the best performance on the Scan2Cap validation set in terms of CIDEr (C), and remains highly competitive on other metrics such as BLEU-4, METEOR, and ROUGE-L, demonstrating strong overall effectiveness for 3D captioning.

\begin{table}[h]
\centering
\begin{minipage}{0.54\linewidth}
    \centering
    \caption{Performance comparison on the test set of SQA3D \cite{sqa3d}.}
    \resizebox{\linewidth}{!}{
    \begin{tabular}{lccccccc}
    \toprule
    \multirow{2}{*}{Method} & \multicolumn{6}{c}{Test set} & \multirow{2}{*}{\textbf{Avg.}} \\
    \cmidrule(lr){2-7}
     & What & Is & How & Can & Which & Others & \\ 
    \midrule
    SQA3D~\cite{sqa3d} & 31.60 & 63.80 & 46.00 & 69.50 & 43.90 & 45.30 & 46.60 \\
    3D-VisTA~\cite{3djcg} & 34.80 & 63.30 & 45.40 & 69.80 & 47.20 & 48.10 & 48.50 \\
    LLaVA-Video\cite{llavavideo} & 42.70 & 56.30 & 47.50 & 55.30 & 50.10 & 47.20 & 48.50 \\
    Scene-LLM~\cite{scenellm} & 40.90 & 69.10 & 45.00 & 70.80 & 47.20 & 52.30 & 54.20 \\
    LEO \cite{leo} & -- & -- & -- & -- & -- & -- & 50.00 \\
    ChatScene \cite{chatscene} & 45.40 & 67.00 & 52.00 & 69.50 & 49.90 & 55.00 & 54.60 \\
    LLaVA-3D \cite{llava3d} & -- & -- & -- & -- & -- & -- & 55.60 \\
    Video 3D-LLM~\cite{video3dllm} & 51.10 & 72.40 & 55.50 & 69.80 & {51.30} & 56.00 & 58.60 \\
    \rowcolor{myblue}
    \textbf{3DRS (Ours)} & {54.40} & {75.20} & {57.00} & {72.20} & {49.90} & {59.00} & {60.60} \\
    \bottomrule
    \end{tabular}
    }
    \label{tab:sqa3d}
\end{minipage}%
\hfill
\begin{minipage}{0.44\linewidth}
    \centering
    \caption{{Performance comparison on the validation set of Scan2Cap~\cite{scan2cap}.}}
    \resizebox{\linewidth}{!}{
    \begin{tabular}{lcccc}
    \toprule
    \multirow{2}{*}{Method} & \multicolumn{4}{c}{@0.5} \\
     & C & B-4 & M & R \\
    \midrule
    Scan2Cap~\cite{scan2cap} & 39.08 & 23.32 & 21.97 & 44.48 \\
    3DJCG~\cite{3djcg}& 49.48 & 31.03 & 24.22 & 50.80 \\
    D3Net~\cite{d3net}  & 62.64 & 35.68 & 25.72 & 53.90 \\
    3D-VisTA \cite{3dvista}  & 66.90 & 34.00 & 27.10 & 54.30 \\
    LL3DA \cite{ll3da}  & 65.19 & 36.79 & 25.97 & 55.06 \\
    LEO \cite{leo} & 68.40 & 36.90 & 27.70 & 57.80 \\
    {ChatScene} \cite{chatscene} & {77.19} & 36.34 & {28.01} & {58.12} \\
    LLaVA-3D \cite{llava3d} & 79.21  &41.12 & {30.21} & {63.41} \\
    Video 3D-LLM \cite{video3dllm} & 83.77 & {42.43} & 28.87 & 62.34 \\
    \rowcolor{myblue}
    \textbf{3DRS (Ours)} & {86.11} & 41.63 & 28.97 & 62.29 \\
    \bottomrule
    \end{tabular} 
    }
    \label{tab:scan2cap}
\end{minipage}
\end{table}

\subsection{Ablation Study}

\begin{table*}[t]
\centering
\scriptsize
\setlength{\tabcolsep}{1.2mm}
\caption{Distillation on multiple layers.}
\begin{tabular}{lcccccccccc}
\toprule
Supervision & \multicolumn{2}{c}{ScanRefer} & \multicolumn{2}{c}{Multi3DRefer} & \multicolumn{2}{c}{Scan2Cap} & \multicolumn{2}{c}{ScanQA} & SQA3D \\
\cmidrule(lr){2-3} \cmidrule(lr){4-5} \cmidrule(lr){6-7} \cmidrule(lr){8-9} \cmidrule(lr){10-10}
& Acc@0.25 & Acc@0.5 & F1@0.25 & F1@0.5 & C@0.5 & B-4@0.5 & C & EM & EM \\
\midrule
\rowcolor{myblue} Last layer                              & \textbf{62.9} & \textbf{56.1} & \textbf{60.4} & \textbf{54.9} & \textbf{41.6} & \textbf{86.1} & \textbf{104.8} & \textbf{30.3} & \textbf{60.6} \\
Last layer + last 3rd layer             & 61.5 & 54.8 & 60.1 & \textbf{54.9} & 41.4 & 84.4 & 101.4 & 29.2 & 60.5 \\
Last layer + last 3rd + last 5th layer  & 60.5 & 53.9 & 59.0 & 53.8 & 40.0 & 81.1 & 102.9 & 30.0 & 59.6 \\
\bottomrule
\end{tabular}
\label{tab:multi_layer_distill}
\end{table*}

\begin{table*}[b]
\centering
\scriptsize
\setlength{\tabcolsep}{1.2mm}
\caption{Performance with different distillation losses.}
\begin{tabular}{lcccccccccc}
\toprule
Supervision & \multicolumn{2}{c}{ScanRefer} & \multicolumn{2}{c}{Multi3DRefer} & \multicolumn{2}{c}{Scan2Cap} & \multicolumn{2}{c}{ScanQA} & SQA3D \\
\cmidrule(lr){2-3} \cmidrule(lr){4-5} \cmidrule(lr){6-7} \cmidrule(lr){8-9} \cmidrule(lr){10-10}
& Acc@0.25 & Acc@0.5 & F1@0.25 & F1@0.5 & B-4@0.5 & C@0.5 & C & EM & EM \\
\midrule
Euclidean loss           & \textbf{62.9} & \textbf{56.1} & 60.4 & 54.9 & 41.6 & \textbf{86.1} & \textbf{104.8} & \textbf{30.3} & 60.6 \\
Cosine loss               & 62.2 & 55.5 & \textbf{60.4} & \textbf{55.2} & 41.8 & 85.9 & 104.5 & 30.1 & \textbf{60.7} \\
Cosine + Euclidean           & 62.3 & 55.7 & 60.3 & 55.0 & \textbf{42.1} & 85.8 & 102.7 & 29.7 & 60.2 \\
\bottomrule
\end{tabular}
\label{tab:distill_loss}
\end{table*}

\Cref{tab:multi_layer_distill} shows the effect of applying supervision to different numbers of network layers across multiple 3D scene understanding tasks, including object localization, captioning, and question answering. Supervising only the last layer consistently achieves the best performance on all benchmarks. As more intermediate layers are added for supervision, the results degrade. This suggests that multi-layer supervision may over-constrain geometric features and weaken semantic representations, ultimately hindering downstream performance. Future work may explore more advanced strategies to balance geometric and semantic cues.

\Cref{tab:distill_loss} reports the impact of different distillation loss functions, including euclidean loss, cosine loss, and their combination, across various 3D scene understanding benchmarks. The results show that all loss types yield very similar performance, indicating that the choice of feature distance metric has limited influence in our setting.

\subsection{Qualitative Results}

\begin{figure}
    \centering
    \includegraphics[width=0.85\linewidth]{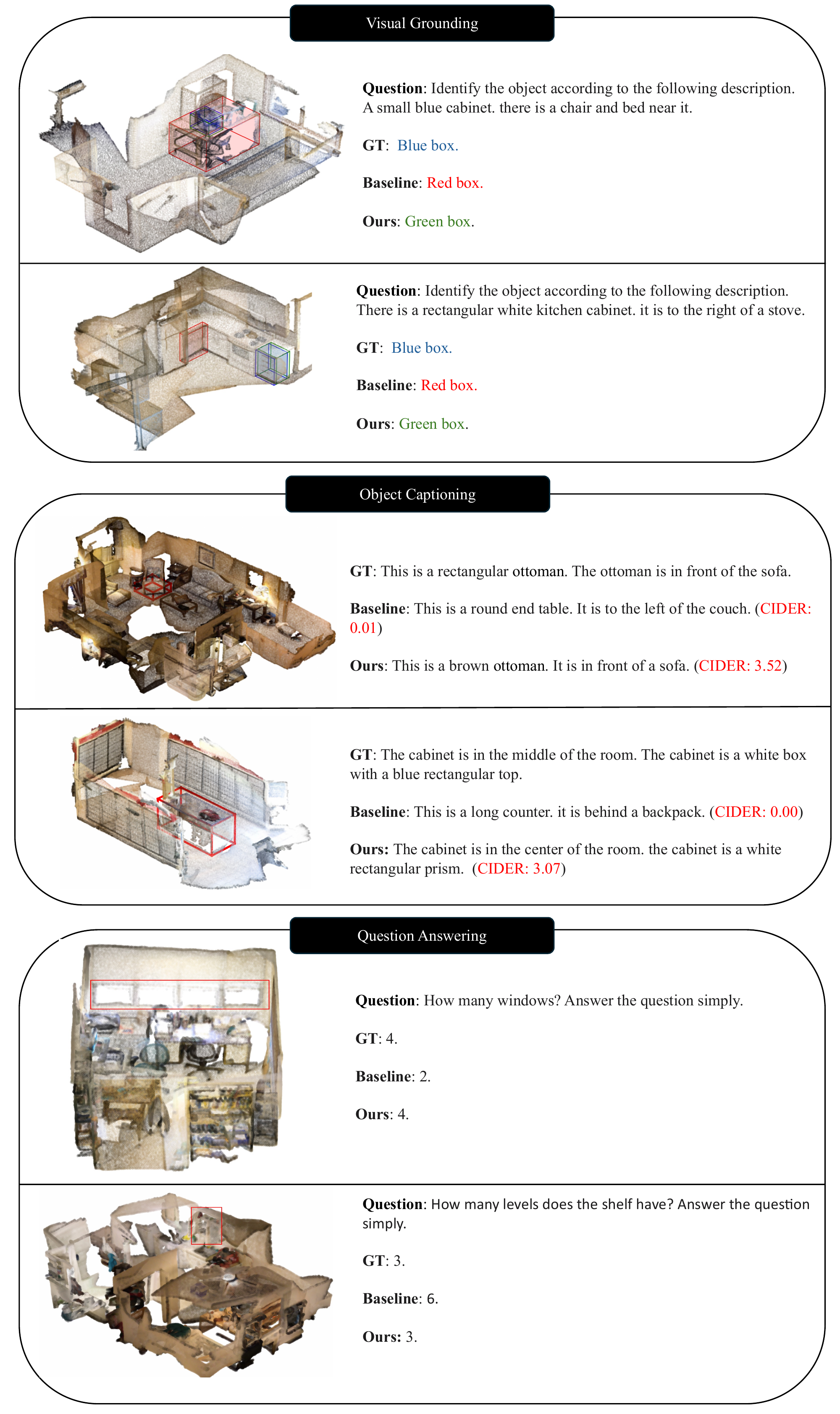}
    \caption{\textbf{Visualization of Results Across Different Tasks.} (a) Visual Grounding: The predicted bounding box closely aligns with the ground truth. (b) Object Captioning: Our method generates accurate captions for each referred object. (c) Question Answering: The model provides precise answers, where we use the red rectangles to indicate the visual cues utilized for each response. Best viewed when zoomed in.}
    \label{fig:visualization_1}
\end{figure}

\textbf{Visualizations}
\label{subsec:vis}
\cref{fig:visualization_1} illustrates qualitative results of our method across three tasks: visual grounding, object captioning, and question answering.

For the visual grounding task (top two rows), the model is required to localize objects within a 3D scene based on natural language descriptions. Each example shows the ground truth bounding box (blue), the result from a baseline method (red), and our prediction (green). In both cases, our method’s predictions match the ground truth more closely than the baseline, demonstrating improved grounding accuracy.

In the object captioning task (middle two rows), the model generates descriptive captions for specific objects in the scene. The captions from the ground truth, the baseline, and our method are shown alongside their corresponding regions. We also report CIDEr scores to measure caption quality. Our approach produces more accurate and detailed descriptions with significantly higher CIDEr scores compared to the baseline.

For the question answering task (bottom two rows), the model answers questions about the scene. Ground truth answers, baseline outputs, and our results are provided for each question. Red rectangles highlight the visual evidence used by our model to generate the answers. Our method provides correct answers that align with the ground truth, whereas the baseline often fails to do so.

Overall, the visualizations demonstrate that our approach consistently outperforms the baseline across all tasks, delivering more accurate grounding, richer object descriptions, and more reliable answers to visual questions.

\begin{figure}
    \centering
    \includegraphics[width=0.85\linewidth]{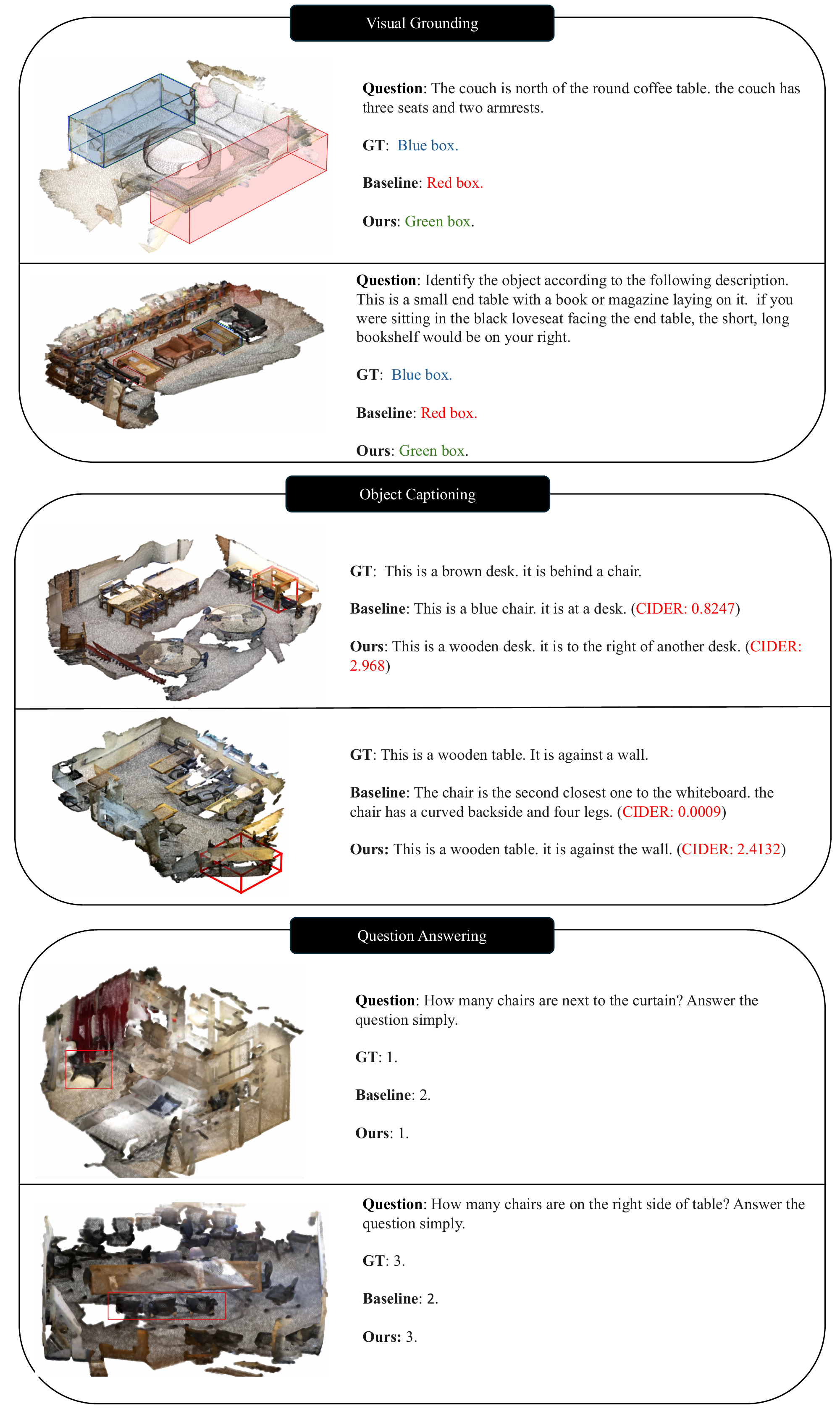}
    \caption{\textbf{Visualization of Results Across Different Tasks.} (a) Visual Grounding: The predicted bounding box closely aligns with the ground truth. (b) Object Captioning: Our method generates accurate captions for each referred object. (c) Question Answering: The model provides precise answers, where we use the red rectangles to indicate the visual cues utilized for each response. Best viewed when zoomed in.}
    \label{fig:visualization_2}
\end{figure}

\begin{figure}
    \centering
    \includegraphics[width=0.85\linewidth]{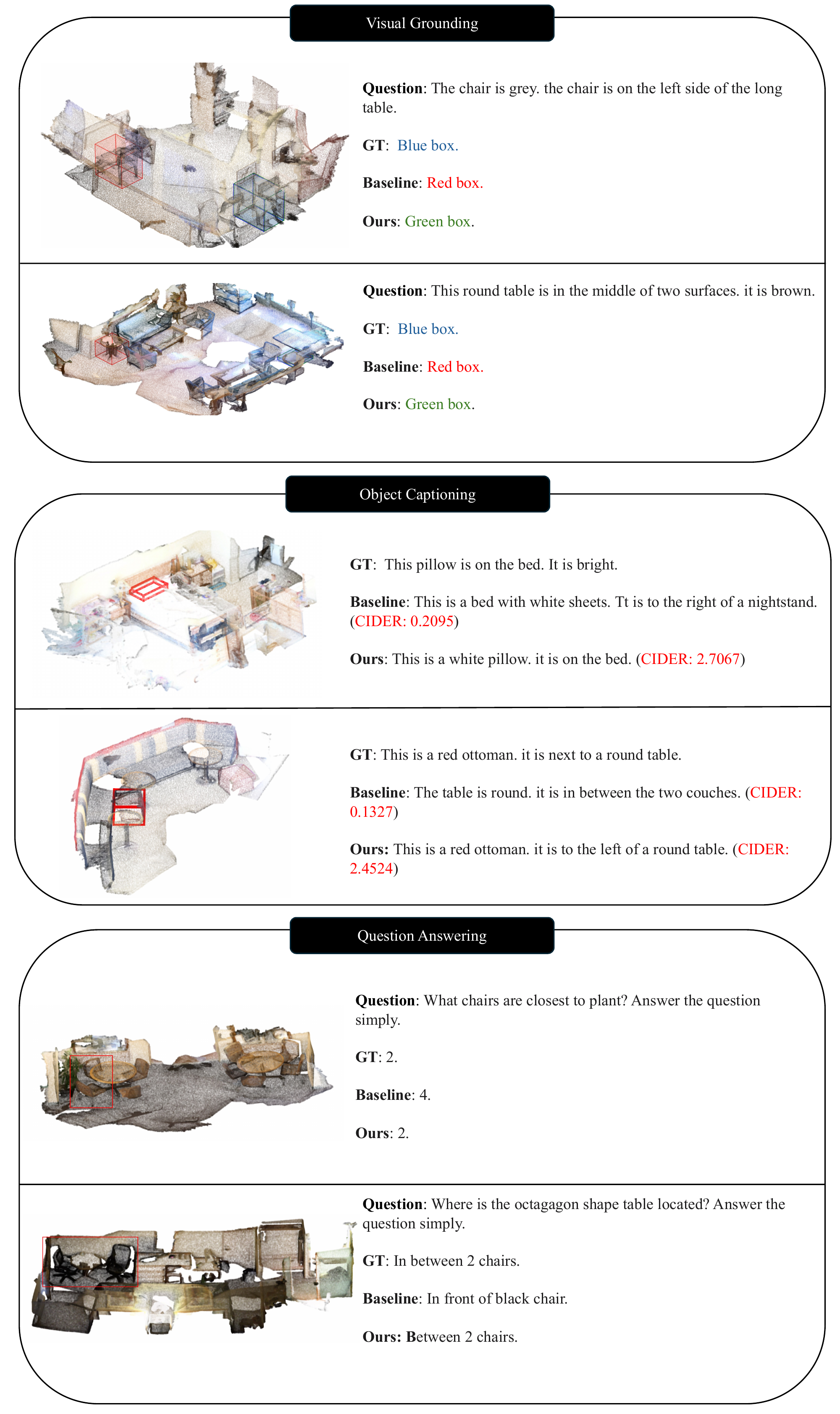}
    \caption{\textbf{Visualization of Results Across Different Tasks.} (a) Visual Grounding: The predicted bounding box closely aligns with the ground truth. (b) Object Captioning: Our method generates accurate captions for each referred object. (c) Question Answering: The model provides precise answers, where we use the red rectangles to indicate the visual cues utilized for each response. Best viewed when zoomed in.}
    \label{fig:visualization_3}
\end{figure}

\cref{fig:visualization_2,fig:visualization_3} provide a visual summary of how our method performs on three challenging 3D scene understanding tasks. These tasks include identifying objects based on language, generating descriptions for specific regions, and answering spatial questions about the scene.

In the visual grounding examples at the top, the model is challenged to find the correct object in a complex 3D environment based on a textual description. The comparison highlights three bounding boxes for each case: blue for the ground truth, red for the baseline, and green for our result. Our predictions consistently align with the intended targets, showing our model’s ability to accurately interpret spatial and semantic cues from language.

The object captioning section in the middle presents how each model describes a highlighted object or area. For each instance, the ground truth, baseline output, and our generated caption are shown, along with their respective CIDEr scores. Our model’s captions are both more precise and more faithful to the scene’s content, as reflected in the higher evaluation scores.

At the bottom, the question answering task demonstrates the model’s reasoning abilities within a 3D environment. The figures show the posed question, the correct answer, the baseline’s response, and our model’s answer. Even for questions that require counting or locating objects, our approach tends to provide accurate answers, often supported by clear visual evidence in the scene.

Altogether, these qualitative results illustrate that our approach delivers more reliable scene understanding across a variety of tasks, outperforming the baseline in both accuracy and descriptive quality.

\subsection{Broader Impacts}
\label{appdix:broaderImpacts}
\textbf{Positive impacts.} The advancement of 3D perception in AI systems holds significant positive societal potential. Enhanced 3D understanding can benefit applications such as assistive robotics for the elderly and disabled, safer autonomous navigation, improved medical imaging, and immersive educational tools. These technologies have the capacity to improve quality of life, boost accessibility, and enable new forms of human-computer interaction.

\textbf{Negative impacts.} However, the adoption of enhanced 3D perception also raises important privacy concerns, especially in surveillance and monitoring contexts where individuals’ activities or environments could be reconstructed and analyzed without their consent. To address these risks, it is crucial to apply robust data anonymization methods---such as blurring faces or removing identifiable features---ensure informed consent from data subjects, enforce strict access controls and data security protocols, and adhere to relevant privacy regulations to protect individual rights.


\newpage

\end{document}